\documentclass{article}

\usepackage{microtype}
\usepackage{graphicx}
\usepackage{subfigure}
\usepackage{caption}
\usepackage{array}
\usepackage{booktabs} 

\usepackage{hyperref}

\usepackage[accepted]{icml2025}

\usepackage{amsmath}
\usepackage{amssymb}
\usepackage{mathtools}
\usepackage{amsthm}
\usepackage{multirow}
\usepackage{bbding}
\usepackage[capitalize,noabbrev]{cleveref}

\theoremstyle{plain}

\theoremstyle{definition}

\theoremstyle{remark}

\usepackage[textsize=tiny]{todonotes}

\icmltitlerunning{Mixup Helps Understanding Multimodal Video Better}

\begin{document}

\twocolumn[
\icmltitle{Mixup Helps Understanding Multimodal Video Better}



\begin{icmlauthorlist}
\icmlauthor{Xiaoyu Ma}{sch1,comp}
\icmlauthor{Ding Ding}{sch1,comp}
\icmlauthor{Hao Chen}{sch1,comp}

\end{icmlauthorlist}

\icmlaffiliation{sch1}{School of Computer Science and Engineering, Southeast University, Nanjing, China}
\icmlaffiliation{comp}{Key Laboratory of New Generation Artificial Intelligence Technology and Its Interdisciplinary
Applications (Southeast University), Ministry of Education, China}

\icmlcorrespondingauthor{Hao Chen}{haochen303@seu.edu.cn}

\icmlkeywords{Balanced Multimodal Learning, Data Augmentation}

\vskip 0.3in
]



\printAffiliationsAndNotice{}  

\begin{abstract}
Multimodal video understanding plays a crucial role in tasks such as action recognition and emotion classification by combining information from different modalities. However, multimodal models are prone to overfitting strong modalities, which can dominate learning and suppress the contributions of weaker ones. To address this challenge, we first propose Multimodal Mixup (MM), which applies the Mixup strategy at the aggregated multimodal feature level to mitigate overfitting by generating virtual feature-label pairs. While MM effectively improves generalization, it treats all modalities uniformly and does not account for modality imbalance during training. Building on MM, we further introduce Balanced Multimodal Mixup (B-MM), which dynamically adjusts the mixing ratios for each modality based on their relative contributions to the learning objective. Extensive experiments on several datasets demonstrate the effectiveness of our methods in improving generalization and multimodal robustness.
\end{abstract}

\section{Introduction}

\begin{figure*}[htbp]
  \centering
  \subfigure[The training and test accuracies of the multimodal model.]{
    \includegraphics[width=0.31\linewidth]{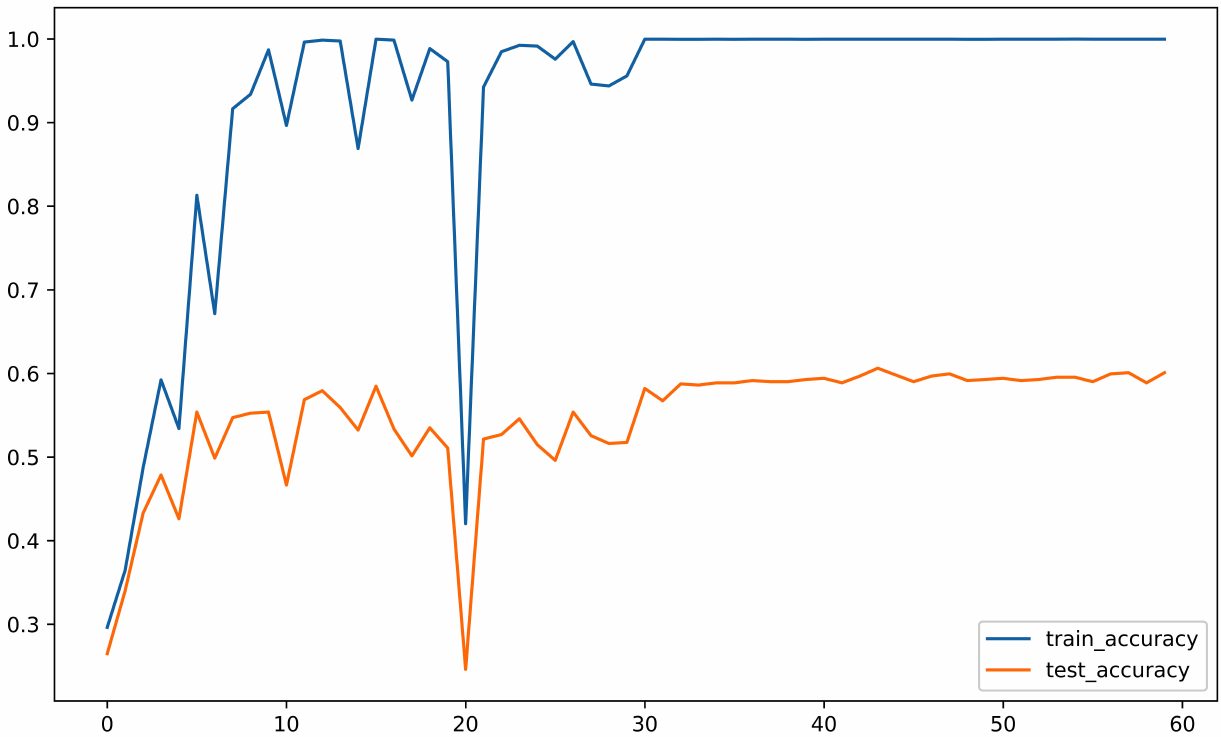}
    \label{fig:acc_base_multi}
  }
  \hfill
  \subfigure[The training accuracies of the multimodal model and its unimodal branches.]{
    \includegraphics[width=0.31\linewidth]{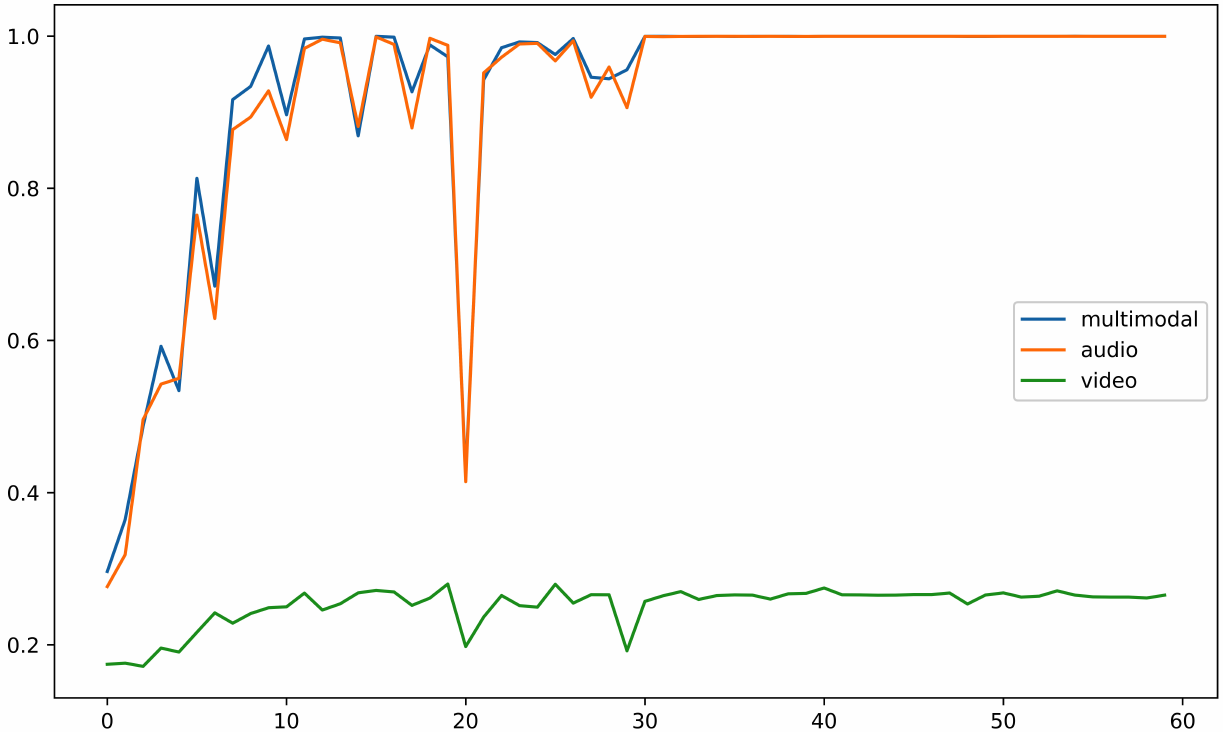}
    \label{fig:acc_base_compare}
  }
  \hfill
  \subfigure[Training and test accuracies of the audio and video branches.]{
    \includegraphics[width=0.31\linewidth]{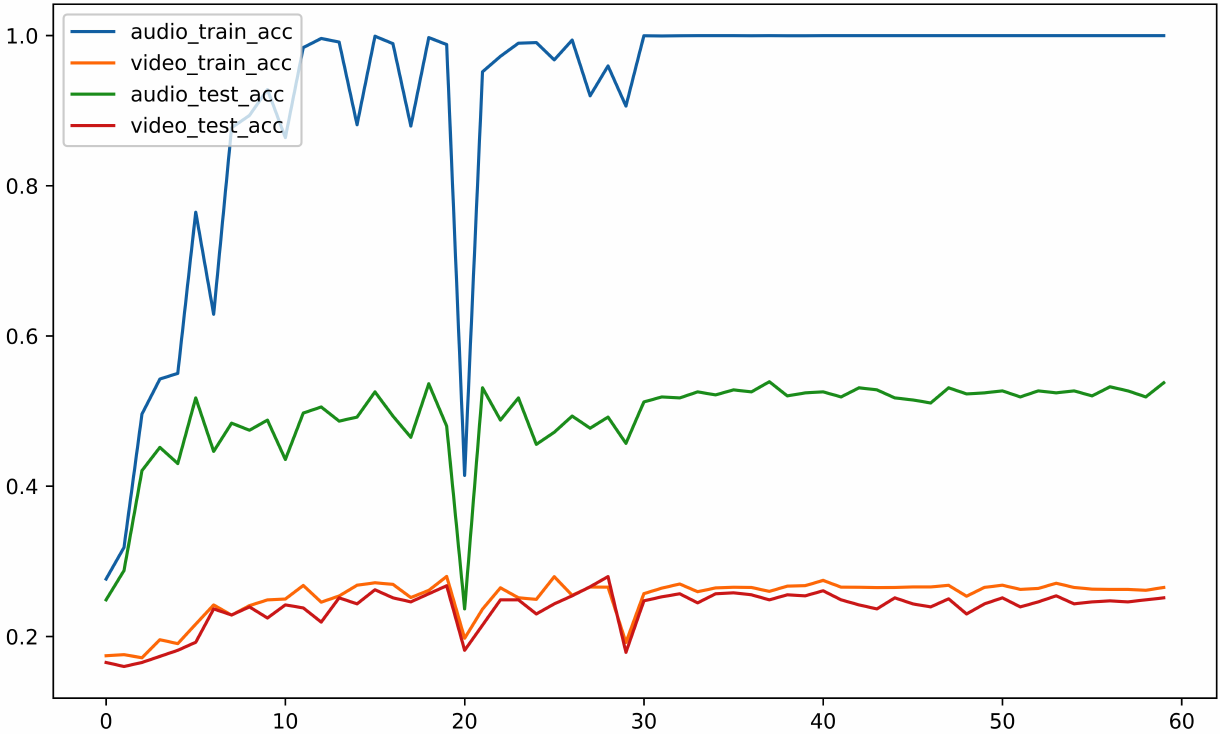}
    \label{fig:acc_train_test}
  }
  \caption{Training and test accuracy curves of the multimodal model and its individual branches during the learning process on the CREMAD dataset.}
  \label{fig: three figs}
\end{figure*}

Multimodal video understanding, as a key research direction in computer vision and multimodal learning \cite{ngiam2011multimodal}, has attracted widespread attention in recent years and plays an important role in tasks such as action recognition \cite{simonyan2014two, feichtenhofer2019slowfast}, event detection \cite{baraldi2017hierarchical, wu2018compressed}, video generation \cite{clark2019adversarial}, and description \cite{rohrbach2016grounding}. With the development of multimodal learning, various modalities—such as visual, audio, and even textual inputs—have been integrated in the hope of enhancing the model’s representational capacity and task performance through modality complementarity \cite{zhu2024vision+}. However, while multimodal inputs provide richer information, they also make models more prone to overfitting specific modalities or spurious correlations in the data. As shown in Figure \ref{fig:acc_base_multi}, the model quickly converges on the training set, yet achieves relatively low accuracy on the test set, demonstrating data memorization and overfitting.

To alleviate multimodal overfitting and improve generalization, data augmentation techniques have become a major focus of research. Among them, mixup \cite{mixup} is a simple and effective data augmentation method that generates virtual samples by linearly interpolating between different samples and their labels, thereby enriching the training distribution. Mixup has demonstrated promising results in single-modality tasks such as image classification and speech recognition. However, its exploration in multimodal scenarios such as video understanding remains relatively limited.

To explore the application of mixup in multimodal video understanding, we first proposed the \textbf{Multimodal Mixup (MM)} method, which applies mixup at the aggregated multimodal feature level in the feature space to mitigate the model’s tendency to overfit the data \cite{zhong2020random}. This approach of directly applying uniform mixing across modalities is simple and achieves certain benefits. However, it fails to fully account for the dynamic contribution differences \cite{hu2022shape} of each modality during training. Due to the modality imbalance \cite{wang2020makes, du2021improving} in multimodal joint learning—where a strong modality that is easier to optimize can quickly converge and dominate the learning process—the model tends to overfit this strong modality (as illustrated in Figure \ref{fig:acc_base_compare}), while other modalities may not have been sufficiently learned. Therefore, applying uniform mixup may not only limit the model’s ability to learn robust cross-modal representations but also exacerbate the risk of overfitting to the strong modality.

To address these issues, we further propose the \textbf{Balanced Multimodal Mixup (B-MM)} method for multimodal video understanding. This method monitors the model’s representational capacity on different modalities during training \cite{peng2022balanced} and dynamically adjusts the degree of mixing for each modality based on this information. In doing so, it guides the model to learn more balanced multimodal representations, effectively reducing reliance on any single modality and enhancing generalization in video understanding tasks. We conduct extensive experiments on several benchmark video understanding datasets, and the results demonstrate that our proposed methods consistently improves performance across different tasks, validating the effectiveness of our methods in preventing overfitting and enhancing multimodal cooperation. We summarize our key contributions as follows:

\begin{itemize}
    \item We introduce a Multimodal Mixup (MM) strategy, which applies the mixup augmentation method to the aggregated multimodal feature space. By creating virtual feature-label pairs, MM enriches the training distribution and effectively mitigates overfitting in multimodal video understanding tasks.
    \item Building upon MM, we propose the Balanced Multimodal Mixup (B-MM) method to address the modality imbalance issue. B-MM dynamically adjusts mixing ratios for each modality based on their relative contributions during training, promoting balanced representation learning and preventing domination by strong modalities.
    \item Comprehensive experiments on benchmark datasets (CREMAD, Kinetic-Sounds, and UCF-101) demonstrate that both MM and B-MM consistently outperform traditional fusion and state-of-the-art balanced multimodal learning methods, significantly enhancing model generalization and robustness.
\end{itemize}

\section{Related Work}
\subsection{Mixup for Data Augmentation}
Mixup was originally proposed as a data augmentation method that performs linear interpolation between two samples and their labels \cite{mixup}. It has achieved remarkable generalization performance in single-modality tasks such as image classification. Building on Mixup, various improved strategies have been proposed in subsequent studies. For example, Manifold Mixup \cite{cao2024survey} performs mixing in the feature embeddings at randomly selected layers rather than only at the raw input level, which helps smooth the decision boundaries and enhance model robustness. MetaMixup \cite{mai2021metamixup} introduces a meta-learning mechanism that adaptively learns the mixing strategy based on validation performance, alleviating overfitting caused by blind sampling.

In the multimodal domain, extending Mixup to fuse different modalities has led to promising research progress. M3ixup \cite{lin2024adapt} employs a two-step (adapting and exploring) strategy along with contrastive learning to mix embeddings from different modalities, thereby enhancing the robustness and representational capacity in the presence of missing modalities. Similarly, M3CoL \cite{kumar2024harnessing} introduces a Mixup-based contrastive loss to better capture cross-modal shared relationships in multimodal contrastive tasks, such as multimodal classification.

\subsection{Balanced Multimodal Learning}
Existing studies have found that different modalities reach sufficient fitting at different speeds during training \cite{wang2020makes}. As a result, when optimizing multimodal models with a unified objective, the strong modality tends to dominate the training process, making the model more prone to overfitting the strong modality while the weak modality remains insufficiently learned \cite{du2021improving}. 

To mitigate this imbalance, prior work has explored adding additional learning objectives for the weak modality \cite{wang2020makes, fan2023pmr, wei2024mmpareto, kontras2024improving, hua2024reconboost} or promoting alignment in optimization rates across modalities \cite{sun2021learning, peng2022balanced, wei2025diagnosing, ma2025improving}. For example, G-Blending \cite{wang2020makes} calculates the optimal mixing mode of modality losses by determining the overfitting status of each modality. PMR \cite{fan2023pmr} introduces a prototype cross-entropy loss for each modality to accelerate the learning of slower modalities. ATF \cite{sun2021learning} dynamically adjusts the learning rates of different modalities based on the unimodal predictive loss. OGM-GE \cite{peng2022balanced} adaptively modulates the gradients of each modality by monitoring discrepancies in their contributions to the learning objective. While these methods promote alignment of optimization rates during training, they often suppress the representational capacity of the strong modality and may be constrained by specific model architectures.

\begin{figure*}[t]
  \centering
  \includegraphics[width=\textwidth]{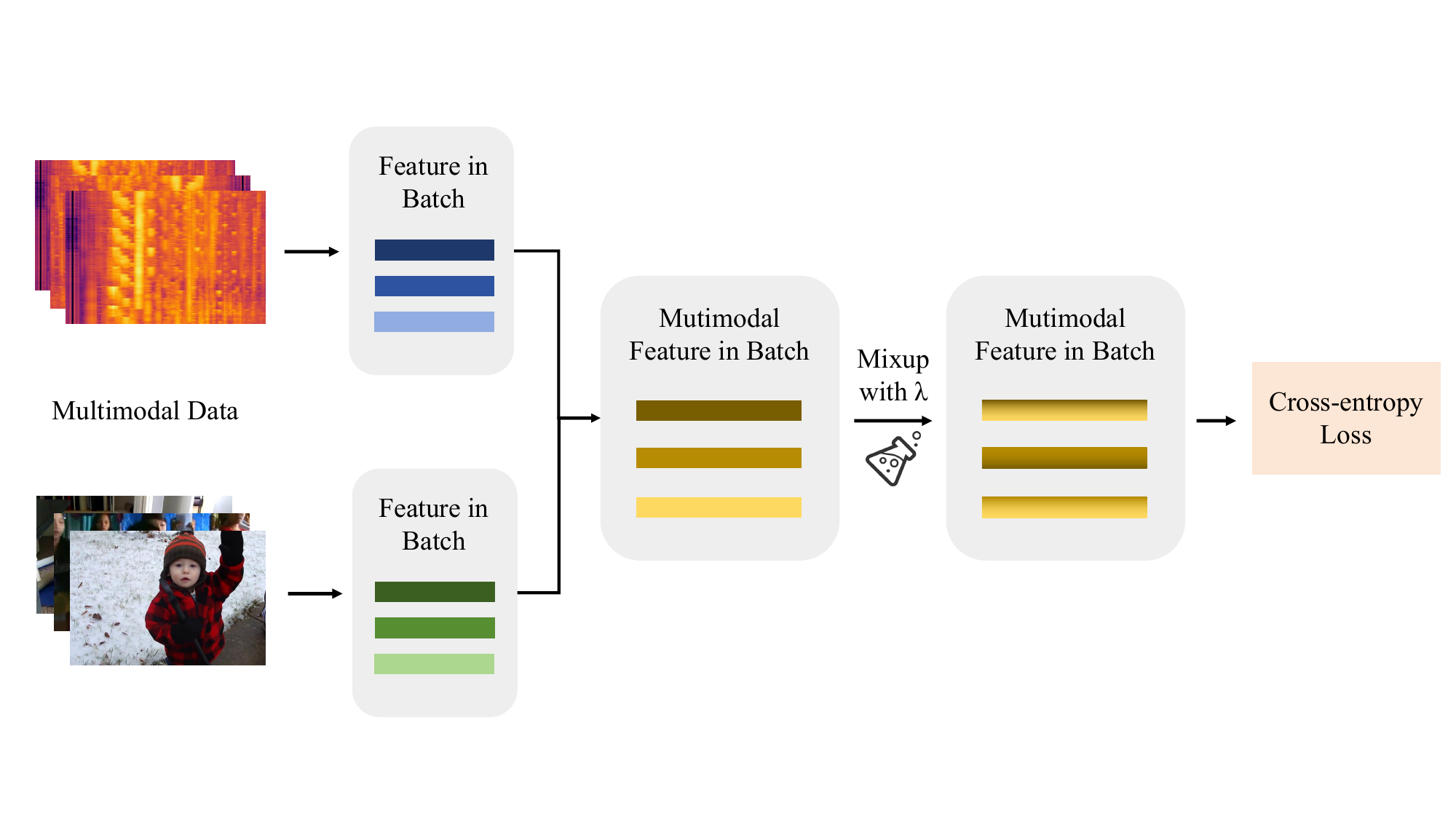}
  \caption{The pipeline of the Multimodal Mixup (MM) method. For each batch, the feature representations of individual modal inputs are first extracted and fused to obtain a multimodal feature representation. The mixup \cite{mixup} method is then applied to the multimodal feature representation with $\lambda$ as the mixing parameter to generate virtual feature-label pairs, which are used for model learning.}
  \label{fig: multimodal mixup}
\end{figure*}

\section{Method}
\subsection{Model Formulation}
This work focuses on multimodal video understanding and the modality imbalance phenomenon within it, with downstream tasks including emotion recognition and action recognition. We primarily consider two input modalities: $m_a$ and $m_v$. The training dataset is denoted as $\mathcal{D} = \{ x_i, y_i \}_{i=1,2,\ldots,N}$, where each $x_i$ consists of multimodal inputs, i.e., $x_i = (x_i^a, x_i^v)$. The label $y_i$ belongs to $\{1, 2, \ldots, M\}$, where $M$ is the number of classes.

We use a multimodal model consisting of two unimodal branches for prediction. Each branch has a unimodal encoder, denoted as $\phi^a$ and $\phi^v$, used to extract features from the corresponding modality of $\boldsymbol{x}$. The encoder outputs are represented as $\boldsymbol{z}^a = \phi^a(\theta^a, \boldsymbol{x}^a)$ and $\boldsymbol{z}^v = \phi^v(\theta^v, \boldsymbol{x}^v)$, where $\theta^a$ and $\theta^v$ are the parameters of the encoders. The results of the two unimodal encoders are fused in some way \cite{concat, dicisionfuse} to obtain the multimodal output. We use Cross-entropy (CE) loss as the loss function and denote it as $\mathcal{L}$.

\subsection{Mixup in Multimodal Learning}

In multimodal supervised learning, we assume that the model receives two modal inputs and obtains the corresponding feature vectors $Z^a$ and $Z^v$ through their respective encoders. Our goal is to identify a function $f \in \mathcal{F}$ that represents the mapping between the feature vectors $(Z^a, Z^v)$ and the target vector $Y$, where these vectors follow the joint distribution $P((Z^a, Z^v), Y)$. We first define a loss function $\mathcal{L}$ that penalizes the discrepancy between the model’s prediction $f(x^a, x^v)$ and the actual target $y$ for a given example $((x^a, x^v), y) \sim P$. However, since the distribution $P$ is unknown, we approximate the expected risk using the empirical risk computed over the available dataset samples:

\begin{align}
    \label{equ:Rf}
  R_\delta(f) &= \int \mathcal{L}(f(x^a,x^v),y)dP_\delta((x^a,x^v),y) \\ &= \frac{1}{n}\sum_{i=1}^n\mathcal{L}(f(x_i^a,x_i^v),y_i),
\end{align}

where $P_\delta$ is the empirical distribution. As noted by \citet{mixup}, learning the function $f$ by minimizing Equation \ref{equ:Rf} can improve computational efficiency. However, when the network has a large number of parameters, a straightforward way to minimize Equation \ref{equ:Rf} is to memorize the training data, which leads to the well-known issue of overfitting.
By observing the multimodal learning process, we find that multimodal models may exhibit data memorization during training. As shown in Figure \ref{fig:acc_base_multi}, the accuracy of the model on the training set increases rapidly to nearly 100\%, while its accuracy on the test set remains significantly lower, indicating poor generalization performance. To improve the model’s performance, we introduce mixup \cite{mixup} into the field of multimodal learning as shown in Figure \ref{fig: multimodal mixup}, where we first get the multimodal feature representations of each sample and then adopt mixup method. Then, virtual feature-target vectors are generated by sampling from a mixed neighborhood distribution:

\begin{equation}
\begin{gathered}
  \tilde{z^a} = \lambda \cdot z_i^a + (1-\lambda) \cdot z^a_j, \quad
  \tilde{z^v} = \lambda \cdot z_i^v + (1-\lambda) \cdot z^v_j, \\
  \tilde{y} = \lambda \cdot y_i + (1-\lambda) \cdot y_j,
\end{gathered}
\label{equ:mixup}
\end{equation}
where $\lambda \sim \text{Beta}(\gamma, \gamma)$, for $\gamma \in (0, +\infty)$.
The parameter $\gamma$ serves as a hyperparameter that controls the strength of data interpolation in mixup. When $\gamma\to 0$, the model is effectively trained using the conventional Empirical Risk Minimization (ERM) method \cite{erm}.

\subsection{Balanced Multimodal-Mixup}

Directly and equally applying the mixup method to all modal inputs is a simple and efficient approach. It can effectively mitigate the model’s tendency to memorize training data and improve performance. However, the performance gains achieved by this strategy are quite limited. This phenomenon arises from the issue of modality imbalance in multimodal learning.

When multimodal inputs are jointly trained with a unified objective, the gradients during backpropagation are determined by the combined contributions of all modalities. This can be expressed as follows:
\begin{equation}
\frac{\partial \mathcal{L} }{f(x^a_i, x^v_i)_c} = \frac{e^{(W[z_i^a; z_i^v]+b)}}{\sum_{k=1}^{M} e^{(W[z_i^a; z_i^v]+b)_k}} - 1_{c=y_i},
\label{equ:imbalance}
\end{equation}

where $W$ and $b$ denote the parameters of the final fully connected layer. From Equation \ref{equ:imbalance}, we can find that when one modality can be optimized and converge quickly, it tends to dominate the overall learning process, preventing other modalities from being sufficiently learned. As shown in Figure \ref{fig:acc_base_compare}, during training, the audio modality can learn quickly, whereas the video modality, which contains richer information, fails to be fully learned throughout the process. Moreover, we calculate the train accuracy and test accuracy for both modalities as shown in Figure \ref{fig:acc_train_test} and find that: The audio modality exhibits a much higher accuracy on the training set compared to the test set, reflecting a learning pattern characterized by data memorization. In contrast, the video modality shows only a small difference in accuracy between the training and test sets, indicating that its overall learning remains insufficient.

\begin{figure*}[t]
  \centering
  \includegraphics[width=\linewidth]{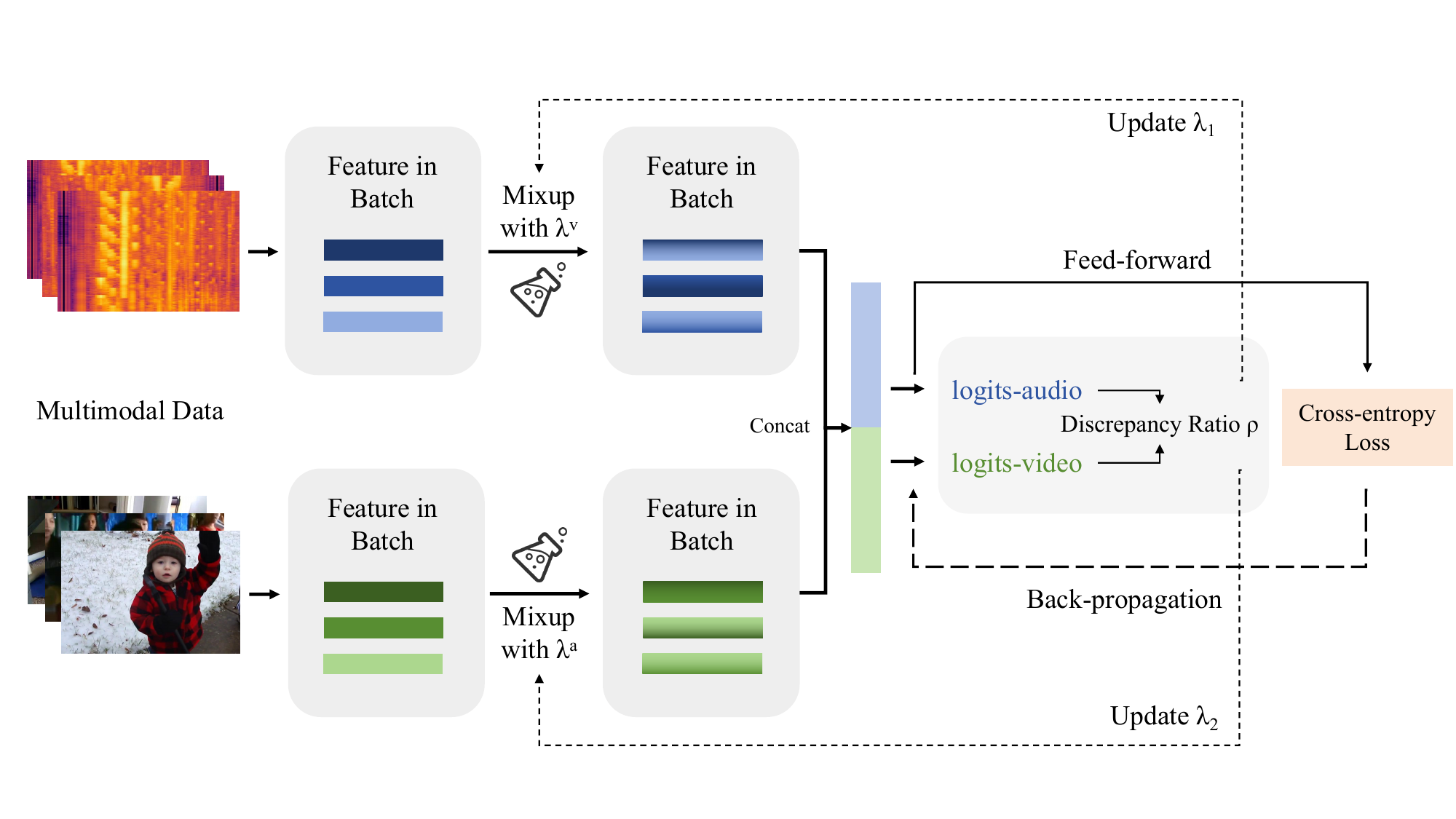}
  \caption{The pipeline of Balanced Multimodal Mixup (B-MM) method. Similar to the MM method, we first get the feature representations of individual modal inputs. Then we apply the mixup method to the unimodal features according to the parameters $\lambda^a$ and $\lambda^v$. After each epoch, the two parameters will update according to the discrepancy ratio $\rho$ \cite{peng2022balanced} of modalities.}
  \label{fig: balanced multimodal mixup}
\end{figure*}

From this, we observe that in the multimodal learning process, due to the modality imbalance problem, it is often the strong modality that tends to memorize the data and overfit. In such cases, simply mixing multimodal features may not only limit the model’s ability to learn robust cross-modal representations but also exacerbate the risk of overfitting to the strong modality. Therefore, we use the modality differences observed during training as a reference to determine which modality should undergo mixup and to what degree, as illustrated in Figure \ref{fig: balanced multimodal mixup}.

Specifically, we first compute the imbalance factor $\rho$ of the model \cite{peng2022balanced} after each training epoch:
\begin{equation}
\begin{gathered}
  s_i^a = \sum_{k=1}^M1_{k=y_i}\cdot\text{softmax}(W[z_i^a; 0]+\frac{b}{2})_k, \\
  s_i^v = \sum_{k=1}^M1_{k=y_i}\cdot\text{softmax}(W[0; z_i^v]+\frac{b}{2})_k,
\end{gathered}
\label{equ:siav}
\end{equation}

\begin{equation}
\rho^v_e = \frac{\sum_{i\in D}s_i^v}{\sum_{i\in D}s_i^a}.
\label{equ:rho}
\end{equation}
Similar to previous work \cite{peng2022balanced}, we mask one of the modalities to zero and use Equation \ref{equ:siav} to obtain the prediction accuracy of each unimodal branch. However, since our method is not applied at every optimization step, we compute the imbalance factor $\rho_t^v$ by aggregating statistics over the entire training set as Equation \ref{equ:rho} shows. 

By dynamically monitoring the change in $\rho$ in each training round to reflect the contribution differences between the audio and visual modalities, we are able to adaptively adjust the degree of Mixup applied to each modality’s input in the next epoch as follows:

\begin{equation}
\lambda_t^u = 
\begin{cases} 
\tanh(\alpha \cdot \rho_t^u) & \rho_t^u > 1 \\
0 & \text{otherwise}
\end{cases},
\label{equ:lambda}
\end{equation}
where $\alpha$ is a hyperparameter that controls the degree of mixup. We use the $\tanh$ function to regulate $\rho$, ensuring that it is constrained within the range $[0, 1]$ while preserving monotonicity. Specifically, when the performance of one modality is better, we apply the mixup method to the other modality’s input according to the computed $\lambda_t^u$ value, as follows:
\begin{equation}
\begin{gathered}
  \tilde{z^a} =  z_i^a, \\
  \tilde{z^v} = \lambda_t^a \cdot z_i^v + (1-\lambda_t^a) \cdot z^v_j, \\
  \tilde{y} = \lambda_t^a \cdot y_i + (1-\lambda_t^a) \cdot y_j,
\end{gathered}
\label{equ:b-mixup}
\end{equation}
when $\rho_t^a >1$, which means audio is the strong modality.

Since the labels are coupled with the modal inputs, this dynamic Mixup applied to the weak modality enables: (1) the strong modality to encounter novel paired modalities and labels, thereby preventing data memorization and overfitting; and (2) the weak modality to generate neighborhood samples, providing greater learning capacity and reducing suppression by the strong modality. This approach helps mitigate modality imbalance during multimodal joint learning and enhances model performance.

\section{Experiments}
\subsection{Dataset and Experimental Settings}

This subsection describes the datasets and experimental settings used in the subsequent study. The main experiments in this work are conducted on three video understanding datasets: CREMAD \cite{cremad}, Kinetic-Sounds \cite{ks}, and UCF101 \cite{soomro2012ucf101}, corresponding to two downstream tasks—emotion recognition and action recognition. 

\begin{table*}[t]
    \centering
  \caption{Combination and comparison with conventional fusion
methods. Bold indicates that our method brings improvement, where “+MM” indicates the use of the Multimodal Mixup method, and “+B-MM” indicates the use of the Balanced Multimodal Mixup method. The best results are highlighted in \textbf{Bold}, and the second-best results are \underline{Underlined}.}
    \vspace{0.2in}
  \label{tab:conventional}
  \begin{tabular}{
@{}>{\centering\arraybackslash}p{4.5cm}|
>{\centering\arraybackslash}p{3cm}||
>{\centering\arraybackslash}p{3.5cm}||
>{\centering\arraybackslash}p{3cm}
@{}}
    \toprule
    Method & CREMAD (Acc) & Kinetic-Sounds (Acc) & UCF-101 (Acc)\\
    \midrule
    Concatenation & 60.62\% & 48.50\% & 79.09 \% \\
    Summation & 57.80\% & 48.84\% & 77.08 \% \\
    Decision Fusion & 61.83\% & 49.34\% & 77.74 \% \\
    FiLM & 59.68\% & 48.65\% & 78.72 \% \\
    Bi-Gated  & 60.89\% & 49.23\% & 77.82 \% \\
    \midrule
    Concatenation + MM & 64.65\% & 50.89\% & 80.81 \% \\
    Summation + MM & 63.58\% & 50.62\% & 79.88 \% \\
   Decision Fusion + MM & 65.86\% & 51.85\% & 80.25 \% \\
   \midrule
    Concatenation + B-MM & \textbf{69.22\%} & \underline{53.66\%} & \textbf{83.32\%} \\
    Summation + B-MM & 68.15\% & 52.58\% & 81.92 \% \\
   Decision Fusion + B-MM & \underline{68.82\%} & \textbf{ 53.82\%} & \underline{82.47\%} \\
    \bottomrule
  \end{tabular}
\end{table*}

\textbf{CREMAD} is an audio-visual dataset for emotion recognition, consisting of 7,442 video clips performed by 91 actors. The dataset covers six common emotion categories: anger, disgust, fear, happiness, neutral, and sadness. A total of 2,443 raters evaluated the emotion and intensity of each clip using three modalities: audiovisual, video-only, and audio-only. The dataset is randomly split into a training and validation set containing 6,698 samples, and a test set containing 744 samples, with a sample ratio of approximately 9:1 between the training/validation and test sets. In this work, we use video frames and audio as multimodal inputs for video understanding.

\textbf{Kinetic-Sounds} is a dataset derived from the Kinetics \cite{kinetic} dataset. The Kinetics dataset contains 400 human action classes collected from YouTube videos, while Kinetic-Sounds selects 31 action classes that are visually and acoustically distinguishable (e.g., playing musical instruments). Each video is manually annotated for human actions using Mechanical Turk and is trimmed into 10-second clips focusing on the action itself. The dataset includes 14,799 clips for training and 2,594 clips for testing. In this work, we use video frames and audio as multimodal inputs for action recognition.

\textbf{UCF-101} is a widely used action recognition dataset that provides both RGB frames and optical flow data, enabling multimodal video analysis tasks. The dataset contains 101 classes of human daily activities, with each video sample sourced from real-world YouTube videos. According to the original dataset split, it consists of 9,537 training samples and 3,783 test samples. In this work, we use video frames and precomputed optical flow frames as multimodal inputs for action recognition.

\textbf{Experimental Settings}. The experiments involve three modalities: video, audio, and optical flow. For all video modalities, frames are sampled at 1 fps, and image frames are uniformly selected as inputs. For the audio modality, spectrograms are generated using Librosa \cite{mcfee2015librosa} and used as inputs. Features for all three modalities are extracted using a ResNet-18 \cite{resnet} network trained from scratch. During training, the Adam \cite{kingma2014adam} optimizer is used for parameter optimization, with $\beta = (0.9, 0.999)$ and a learning rate set to $5 \times 10^{-5}$. All reported results are averaged over three runs with different random seeds, and all models are trained for 60 epochs with a batch size of 64 on two NVIDIA RTX 3090 GPUs to ensure convergence.

\subsection{Comparison with Conventional Fusion Methods}
To validate the effectiveness of our proposed method, we first combine and compare the proposed Multimodal Mixup (MM) and Balanced Multimodal Mixup (B-MM) methods with several classical multimodal fusion approaches in deep learning. Specifically, these include Concatenation (Concat) \cite{concat}, Summation (Sum), Decision Fusion (DeFu) \cite{dicisionfuse}, FiLM \cite{film}, and Bi-Gated \cite{gated}. Among them, methods such as concatenation and FiLM belong to mid-level fusion strategies, while decision fusion represents a late fusion strategy. The results are shown in Table \ref{tab:conventional}. 

We apply Multimodal Mixup and Balanced Multimodal Mixup in combination with concatenation as the representative fusion method for our approaches. It is evident that Multimodal Mixup serves as an effective data augmentation strategy, significantly improving model performance across different datasets and already outperforming other traditional fusion strategies. Furthermore, by incorporating modality imbalance as a reference factor and designing the Balanced Multimodal Mixup method, the model’s performance is further enhanced, with improvements of 8.60\%, 5.16\%, and 4.23\% on CREMAD, Kinetic-Sounds, and UCF-101, respectively. To further demonstrate the generalizability of our method, we combine data mixing with summation and decision fusion, achieving substantial improvements on two datasets.

\begin{figure*}[t]
  \centering
  \par
  \subfigure[Baseline - Audio - Train]{
    \includegraphics[width=0.25\linewidth]{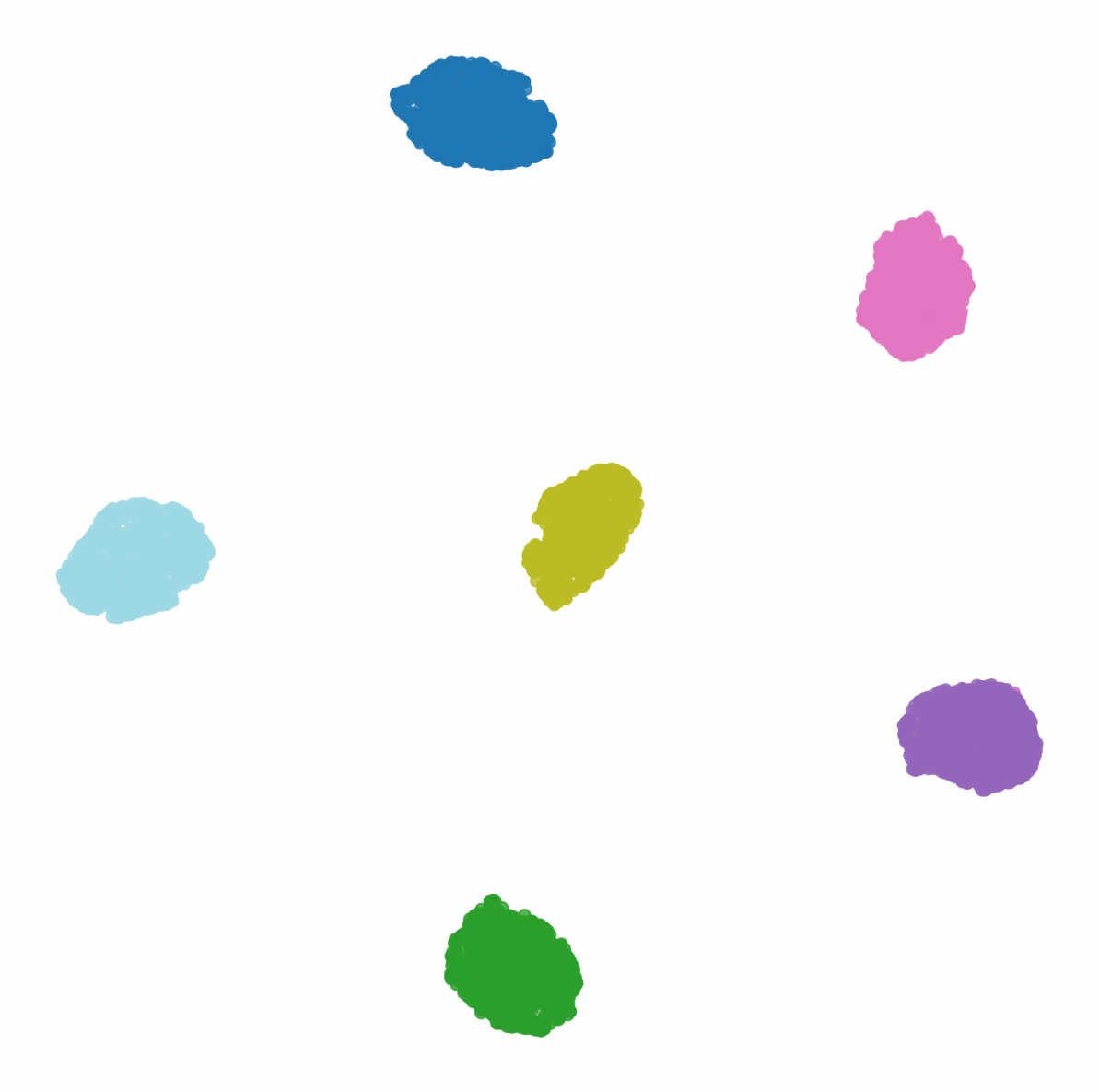}
    \label{fig:class1_1}
  }
  \hfill
  \subfigure[Baseline - Video - Train]{
    \includegraphics[width=0.25\linewidth]{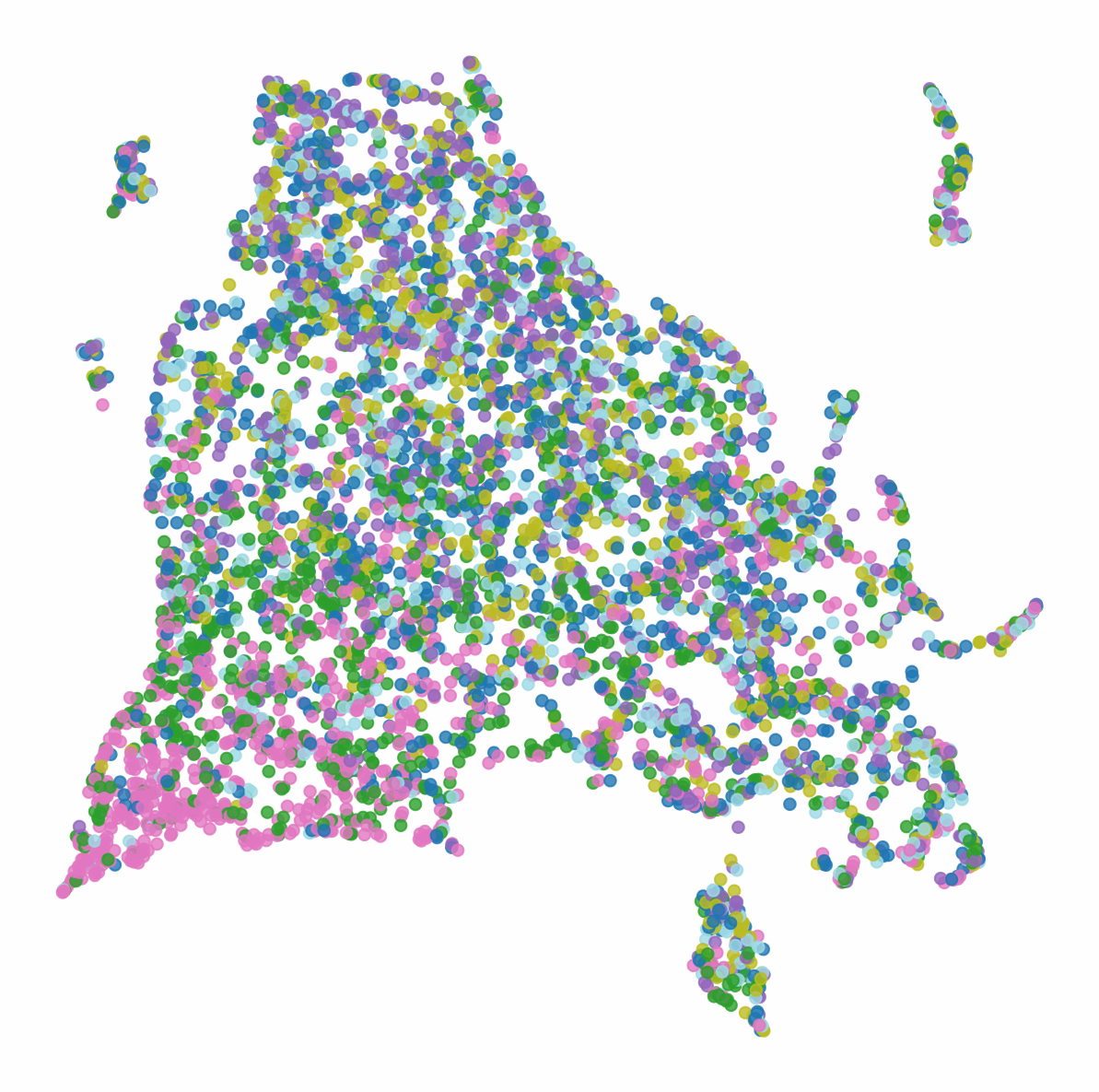}
    \label{fig:class1_2}
  }
  \hfill
  \subfigure[Baseline - Multimodal - Train]{
    \includegraphics[width=0.25\linewidth]{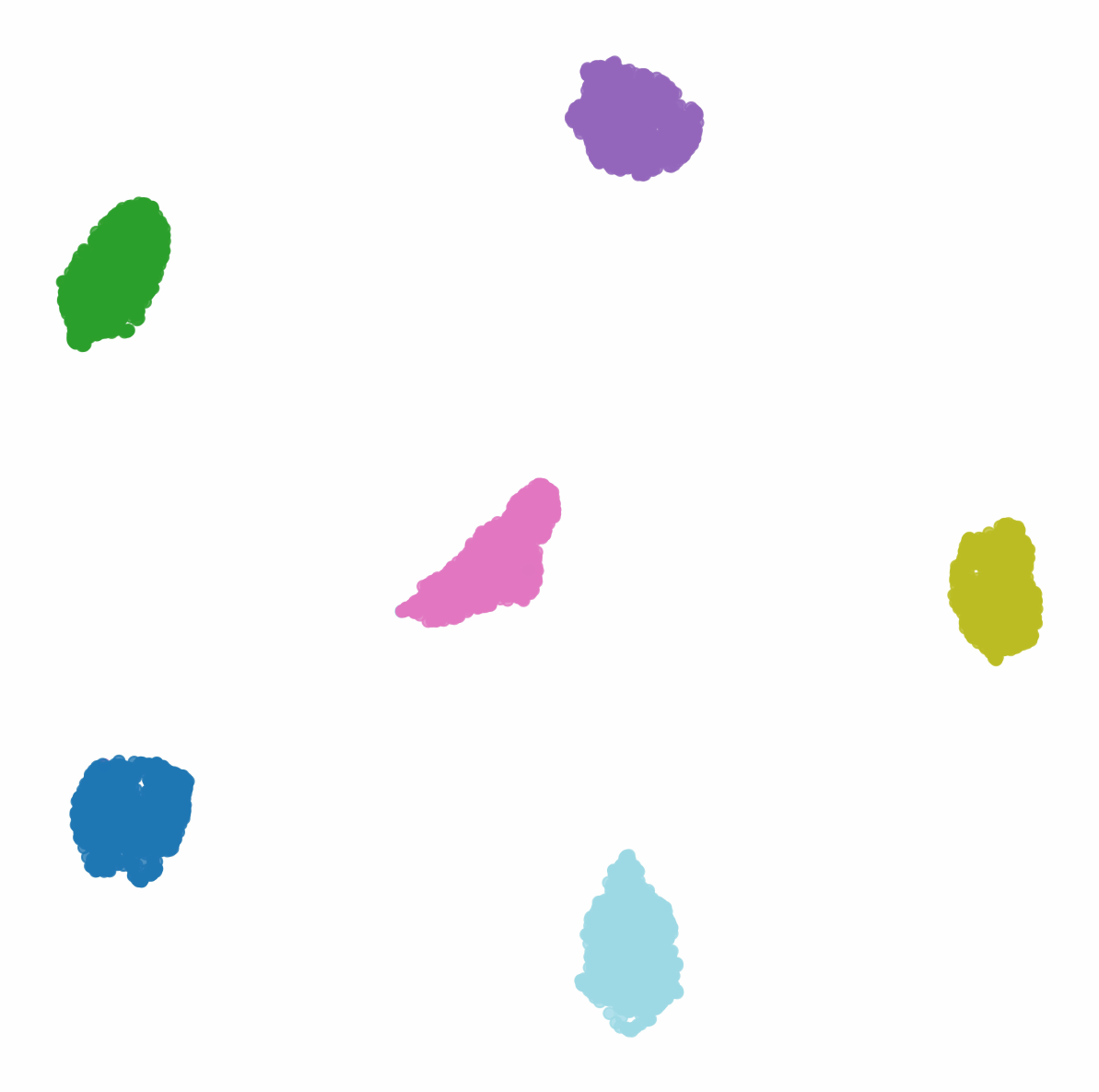}
    \label{fig:class1_3}
  }

  \par
  \subfigure[Baseline - Audio - Test]{
    \includegraphics[width=0.25\linewidth]{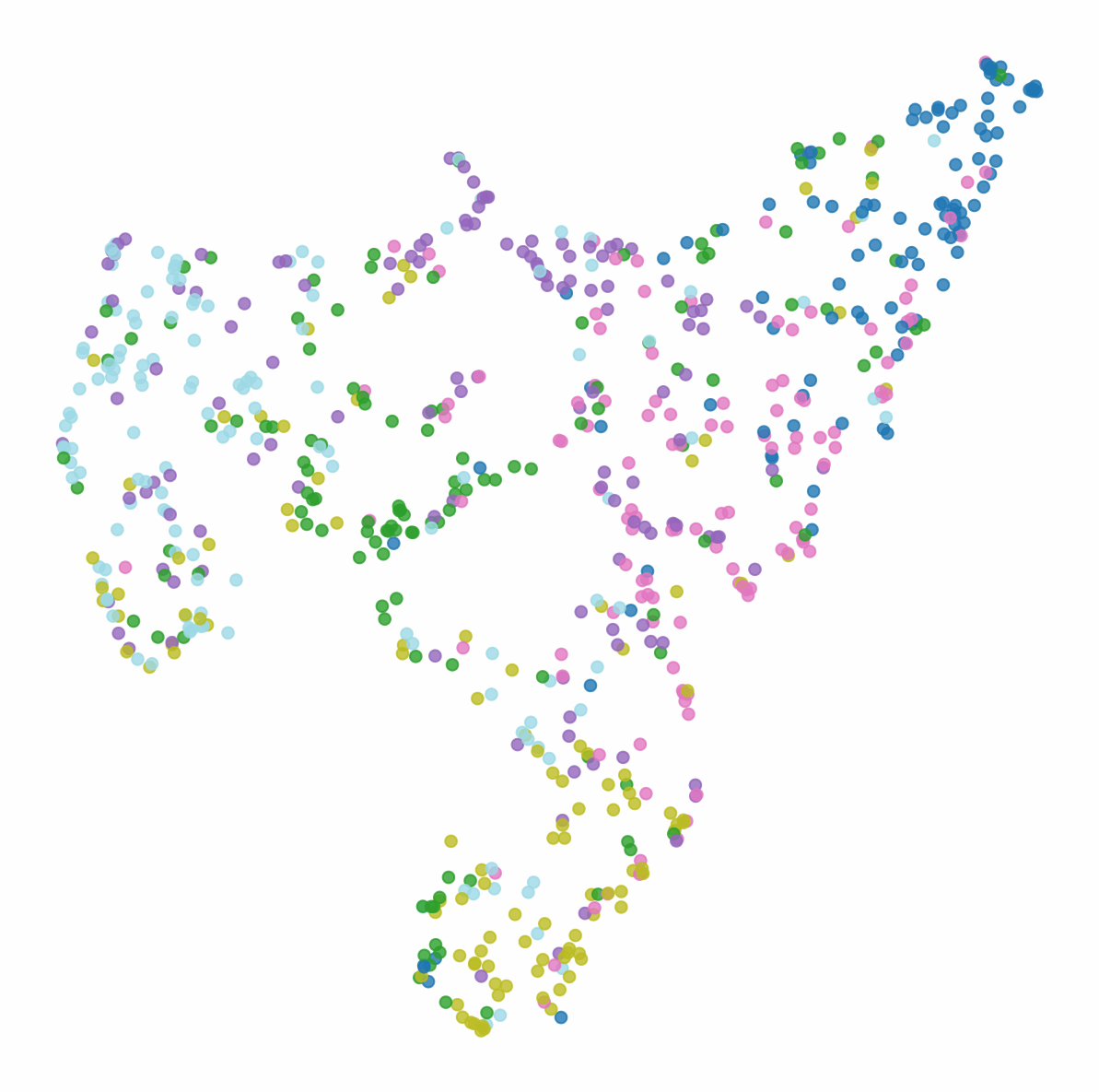}
    \label{fig:class2_1}
  }
  \hfill
  \subfigure[Baseline - Video - Test]{
    \includegraphics[width=0.25\linewidth]{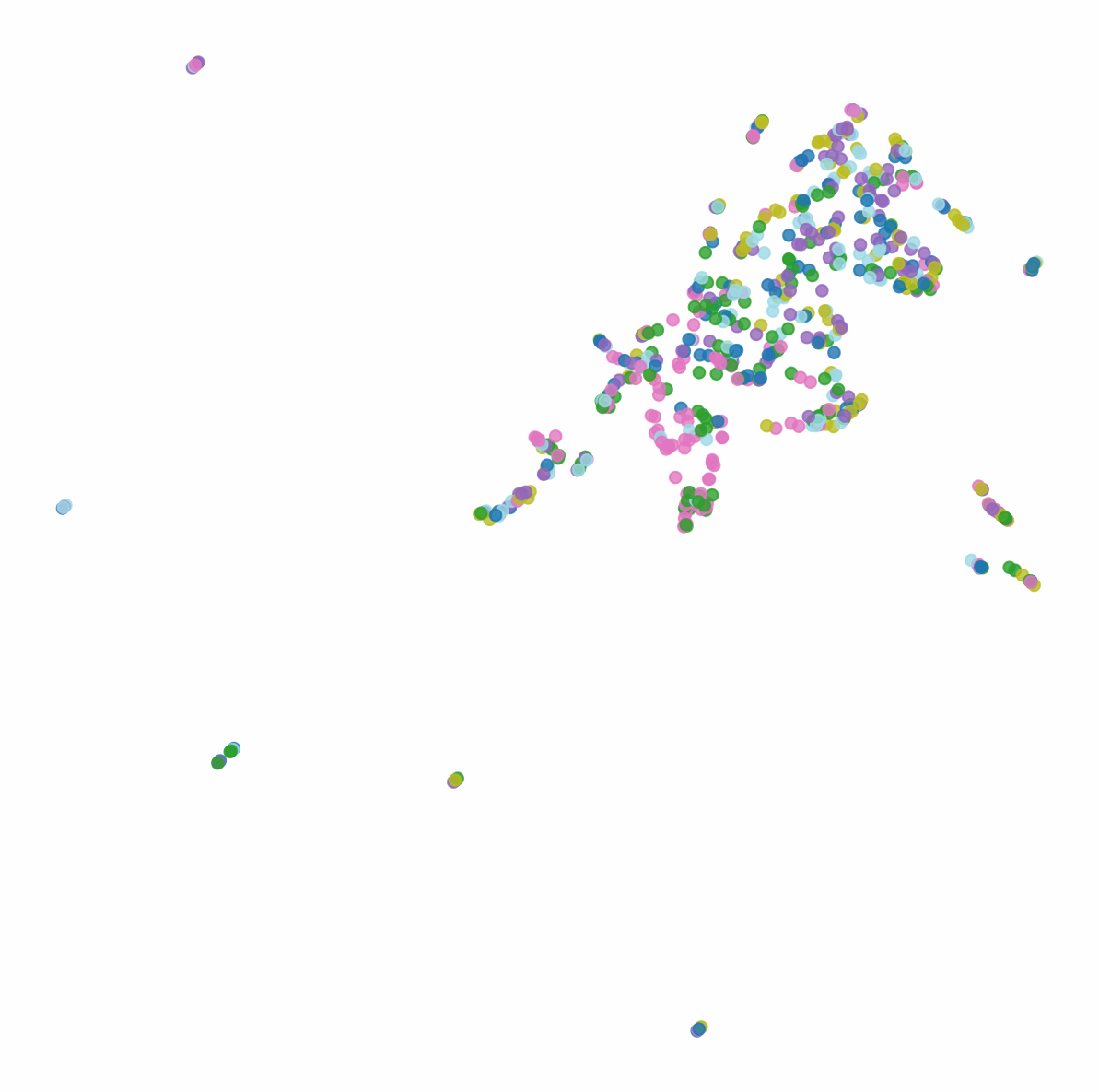}
    \label{fig:class2_2}
  }
  \hfill
  \subfigure[Baseline - Multimodal - Test]{
    \includegraphics[width=0.25\linewidth]{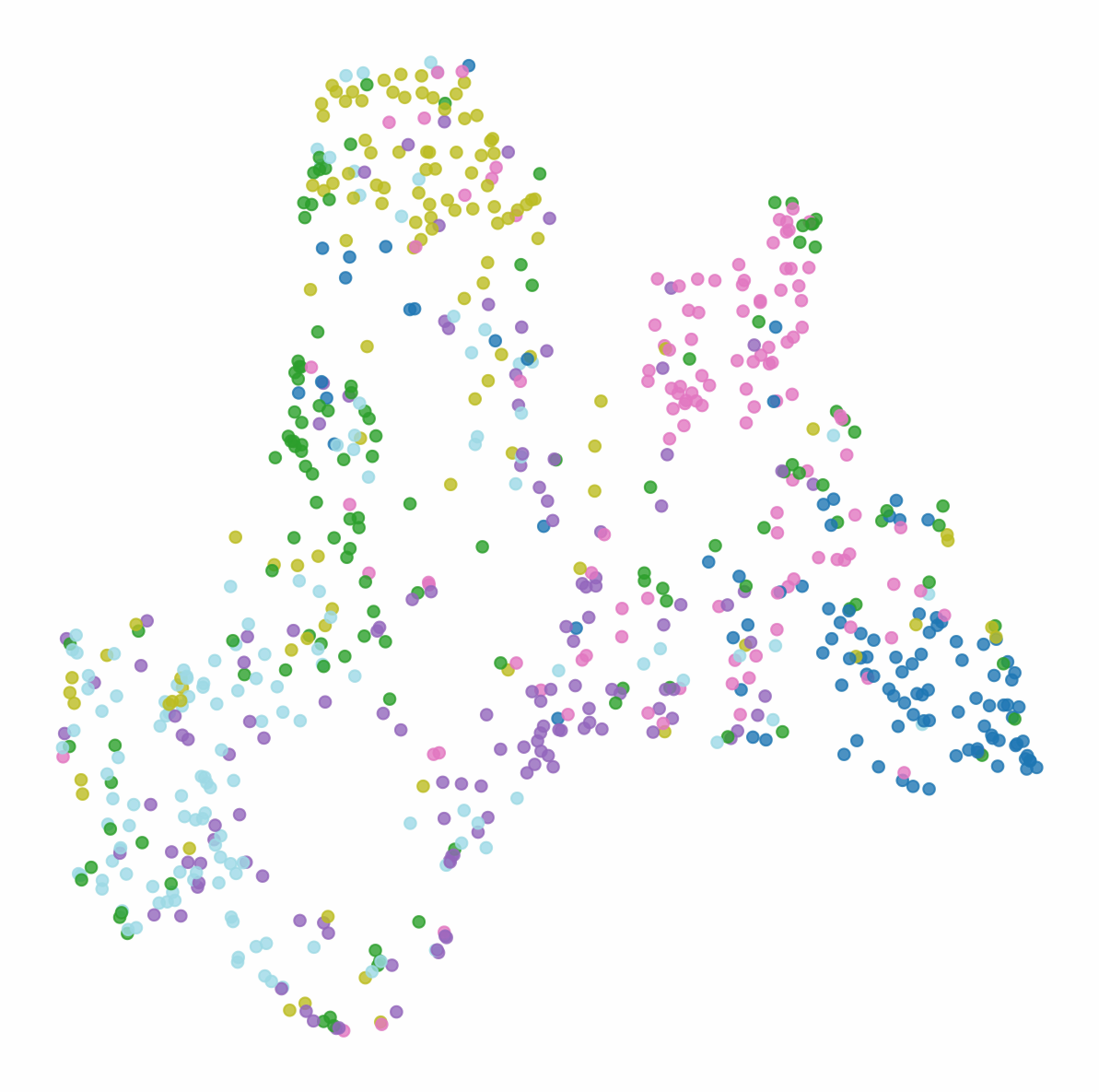}
    \label{fig:class2_3}
  }
  \caption{UMAP visualizations of feature representations from the \textbf{Baseline} models on the training and test sets. Within each configuration, visualizations for audio, video, and multimodal features are included. Different colors indicate different classes.}
  \label{fig:umap base}
\end{figure*}

\begin{figure*}[t]
  \centering
  \par
  \subfigure[B-MM - Audio - Train]{
    \includegraphics[width=0.25\linewidth]{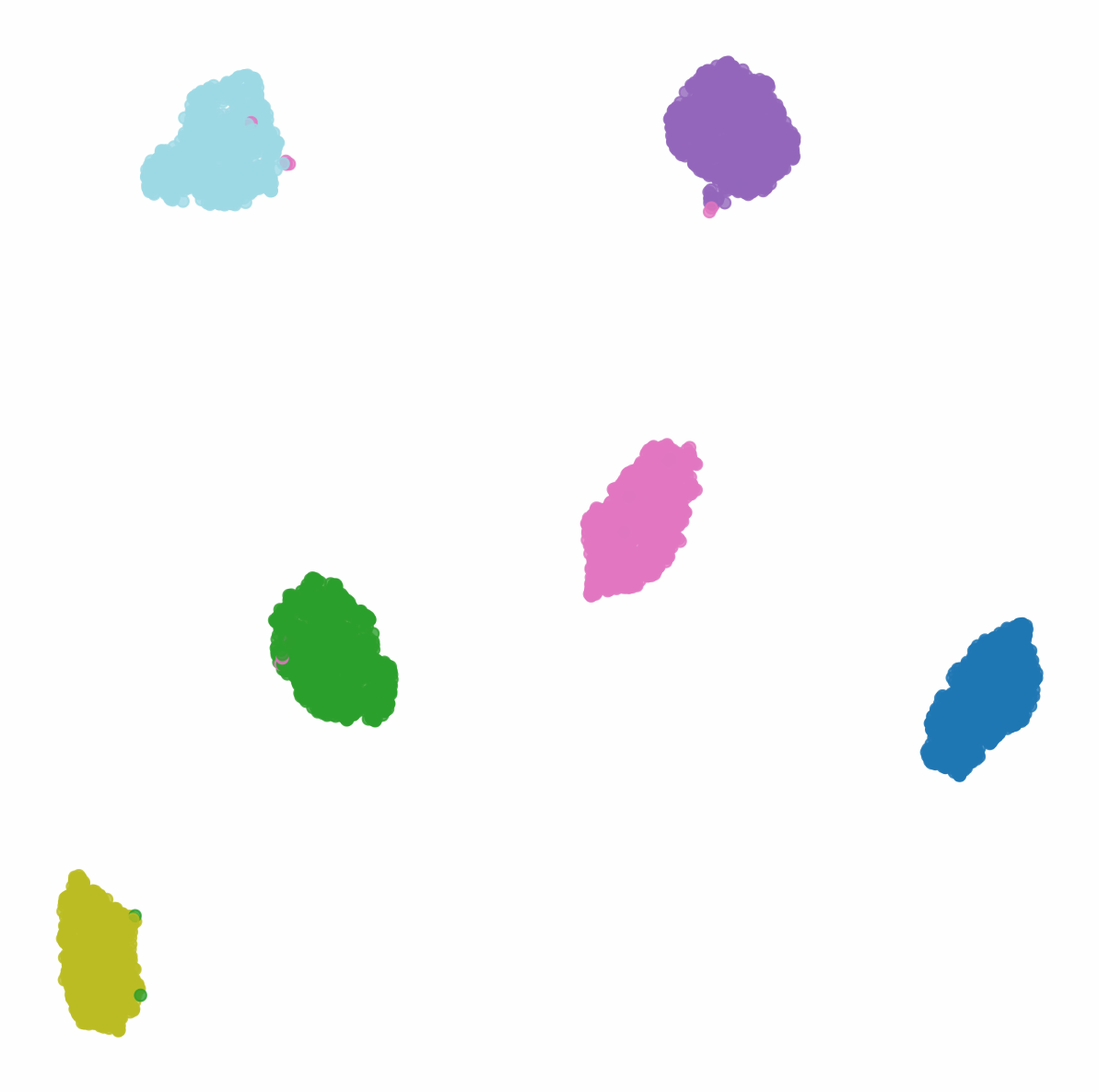}
    \label{fig:class3_1}
  }
  \hfill
  \subfigure[B-MM - Video - Train]{
    \includegraphics[width=0.25\linewidth]{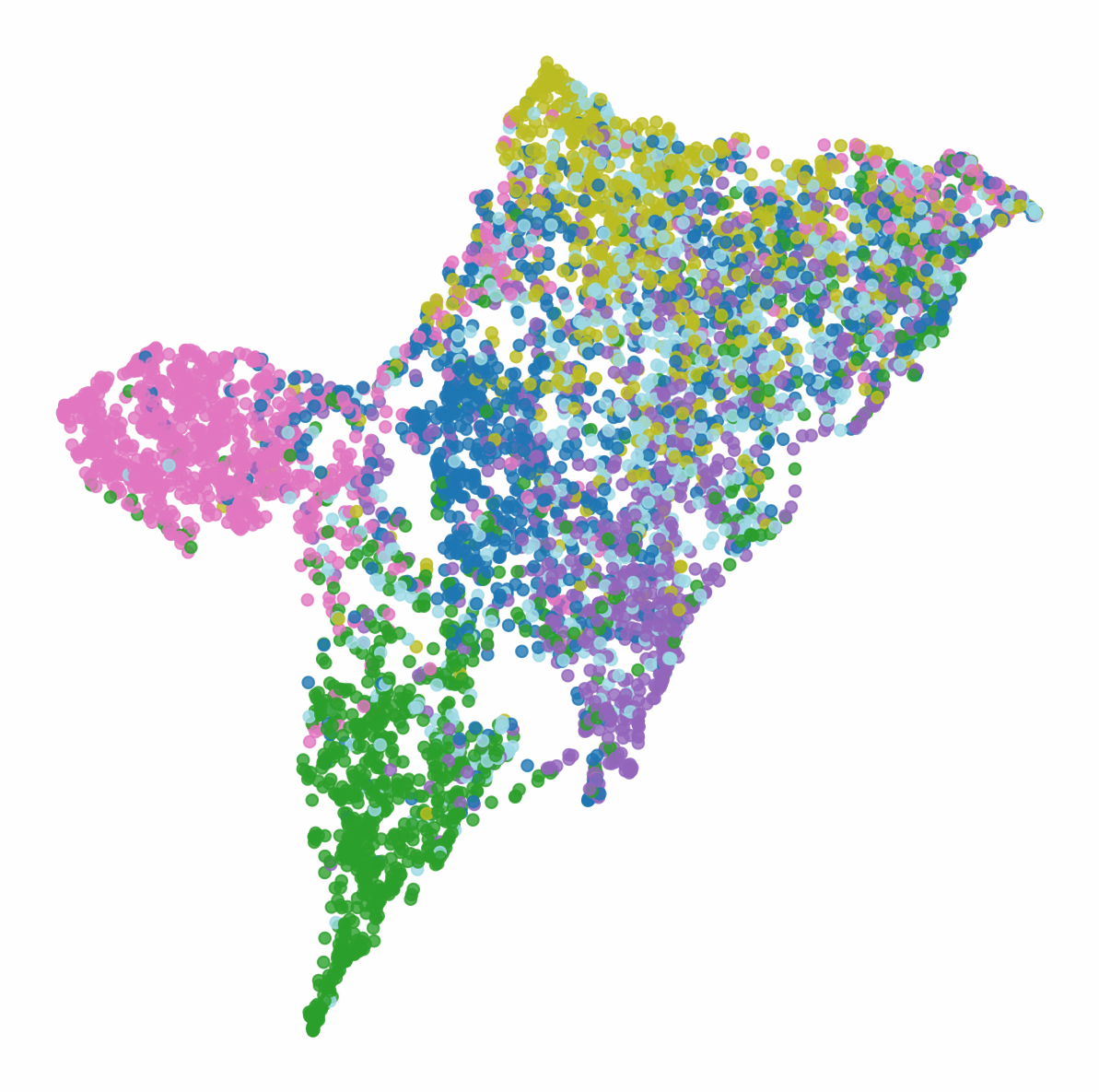}
    \label{fig:class3_2}
  }
  \hfill
  \subfigure[B-MM - Multimodal - Train]{
    \includegraphics[width=0.25\linewidth]{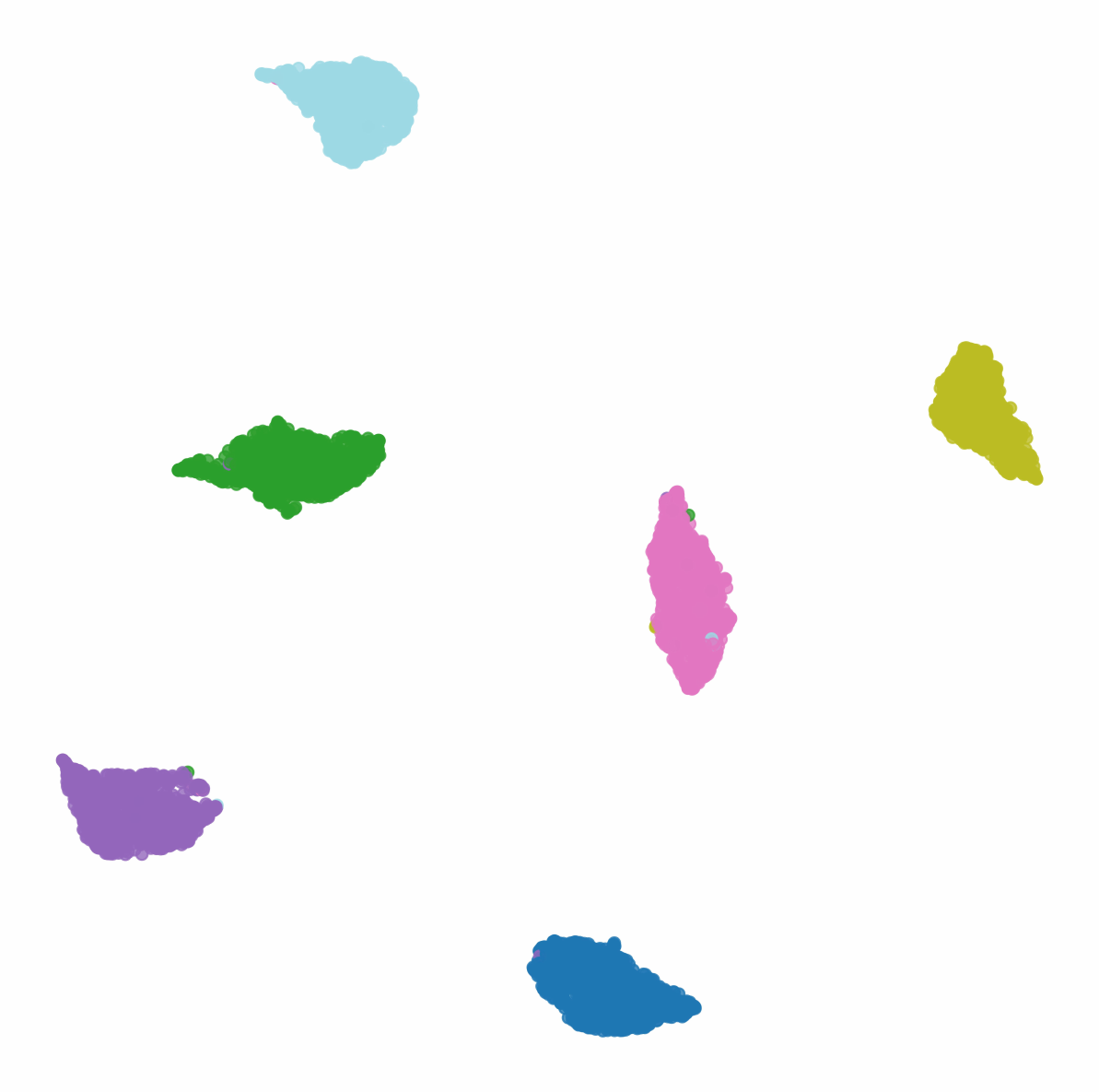}
    \label{fig:class3_3}
  }

  \par
  \subfigure[B-MM - Audio - Test]{
    \includegraphics[width=0.25\linewidth]{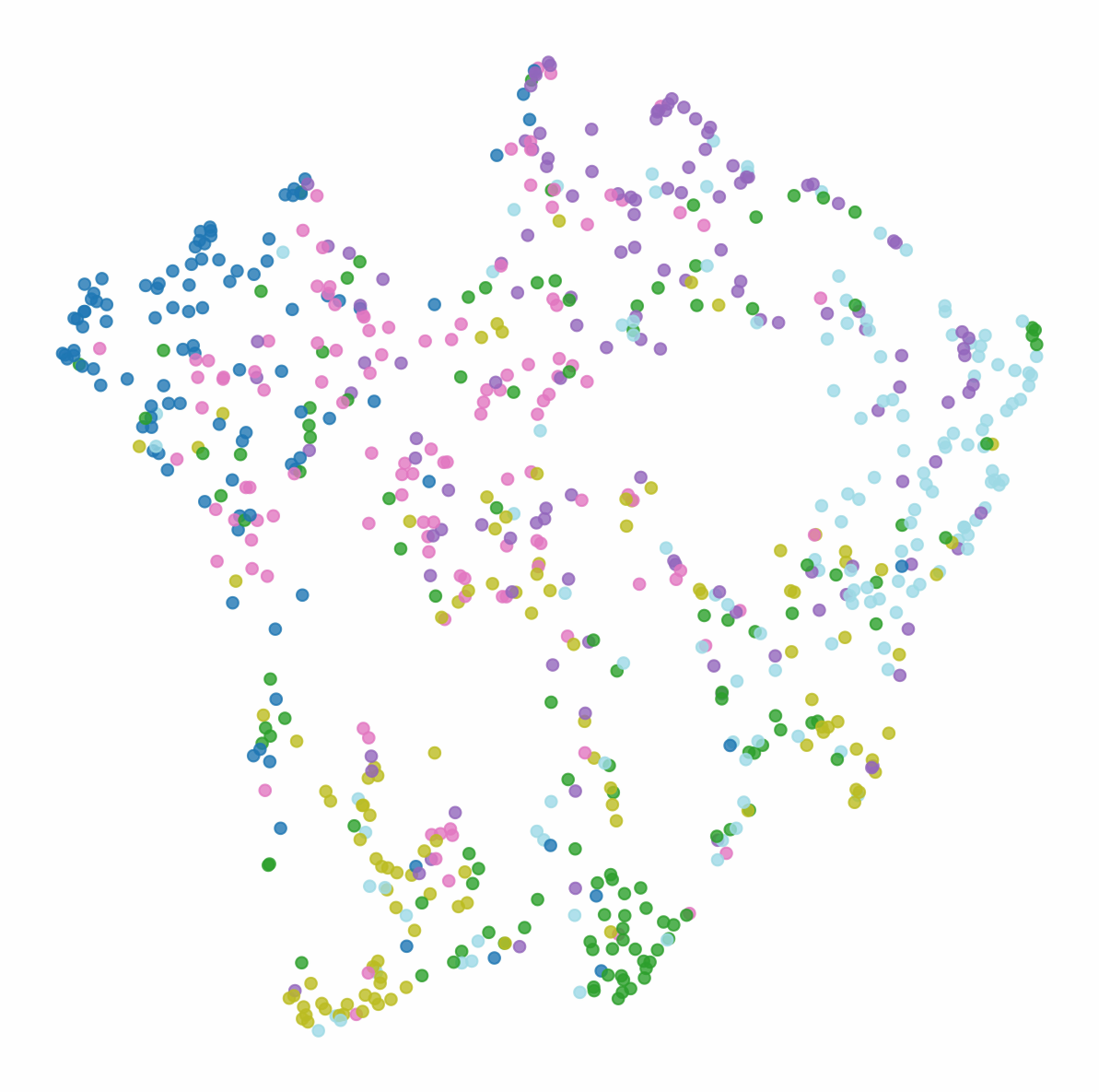}
    \label{fig:class4_1}
  }
  \hfill
  \subfigure[B-MM - Video - Test]{
    \includegraphics[width=0.25\linewidth]{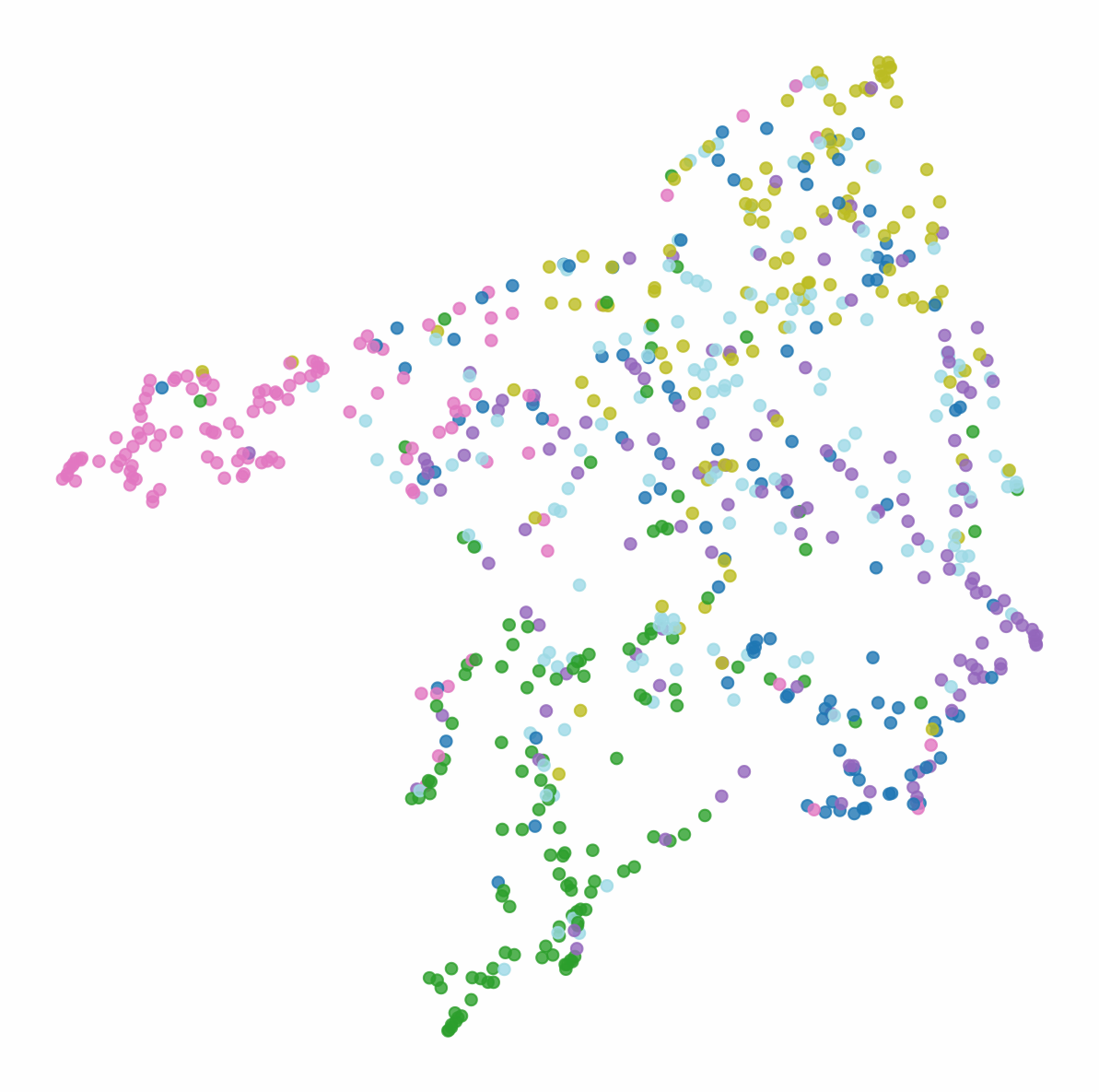}
    \label{fig:class4_2}
  }
  \hfill
  \subfigure[B-MM - Multimodal - Test]{
    \includegraphics[width=0.25\linewidth]{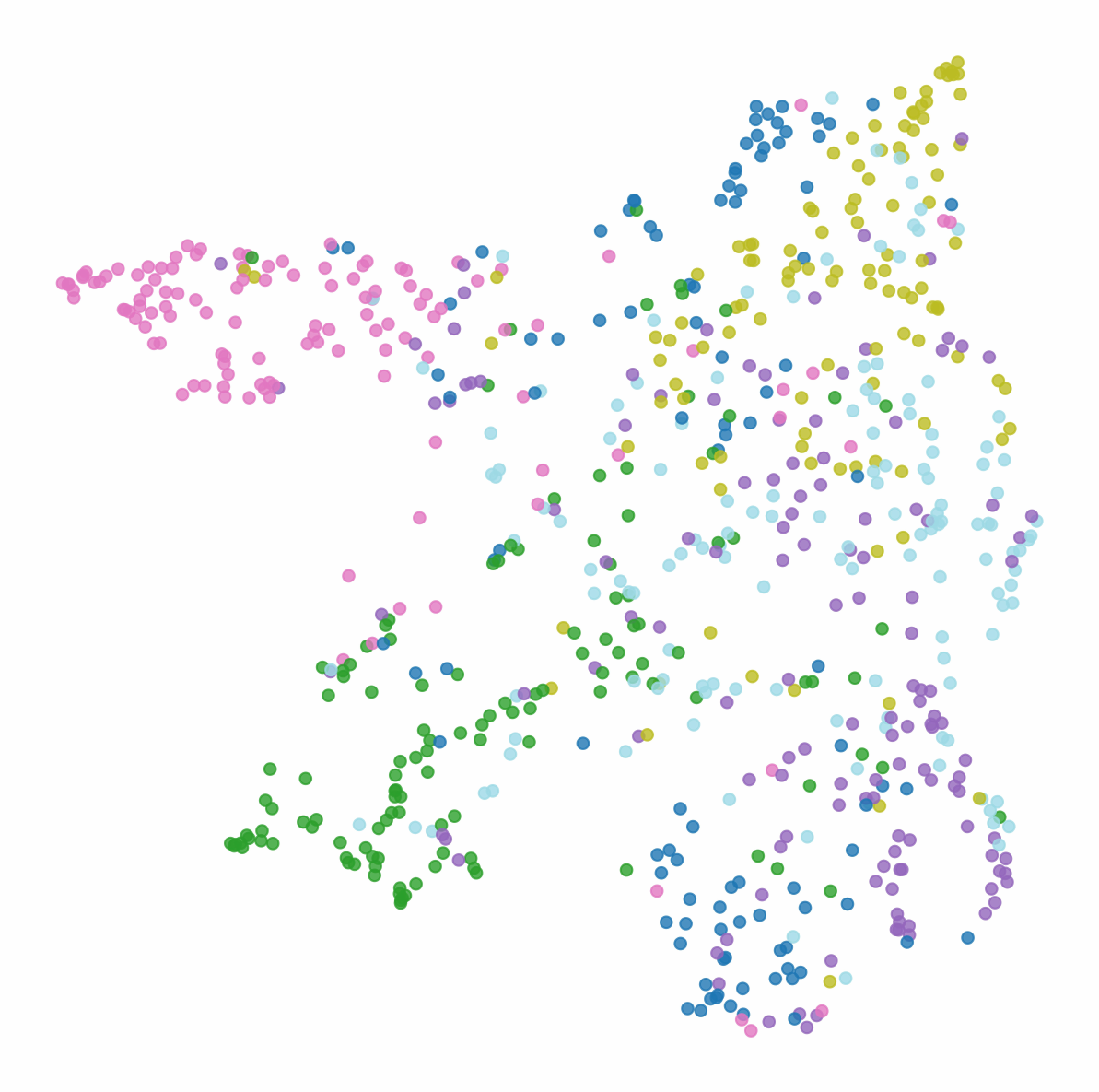}
    \label{fig:class4_3}
  }

  \caption{UMAP visualizations of feature representations from the \textbf{B-MM} models on the training and test sets. Within each configuration, visualizations for audio, video, and multimodal features are provided. Different colors indicate different classes.}
  \label{fig:umap bmm}
\end{figure*}

To provide a more intuitive comparison of the model’s ability to represent data from each modality before and after applying the BMM method, we perform dimensionality reduction and visualization of the multimodal feature outputs as well as individual unimodal features using UMAP \cite{umap}, as shown in Figure \ref{fig:umap base} and Figure \ref{fig:umap bmm}.

\begin{table}[t]
  \caption{Comparison with other imbalanced multimodal learning
methods. All modulation strategies are applied to the baseline,
using Concatenation as the fusion method. The best results are highlighted in \textbf{Bold}, and the second-best results are \underline{Underlined}.}
  \label{tab:bml}
  \begin{tabular}{
  @{}>{\centering\arraybackslash}p{1.6cm}
>{\centering\arraybackslash}p{1.7cm}
>{\centering\arraybackslash}p{2.2cm}
>{\centering\arraybackslash}p{1.5cm}
@{}}
    \toprule
    Method & CREMAD & Kinetic-Sounds& UCF-101\\
    \midrule
    Concat & 60.62\% & 48.50\% & 79.09\% \\
    \midrule
    + G-Blend & 67.34\% & 51.46\% & 82.21\% \\
    + UMT & 65.46\% & 50.31\% & 82.10\% \\
    + PMR & 67.47\% & 51.89\% & 82.34\% \\
    + OGM & \underline{68.55\%} & 51.23\% & \underline{82.55\%} \\
    + Greedy & 66.13\% & \underline{52.43\%} & 81.95\% \\
    + ATF & 64.25\% & 51.54\% & 82.16\% \\
    \midrule
    + MM & 64.65\% & 50.89\% & 80.81\% \\
    + B-MM & \textbf{69.22\%} & \textbf{53.66\%} & \textbf{83.32\%} \\
    \bottomrule
  \end{tabular}
\end{table}

From the visualizations, we observe that when using the baseline multimodal model, the model almost completely memorizes the data from the audio modality (Fig. \ref{fig:class1_1}), while the learning performance for the video modality is extremely poor (Fig. \ref{fig:class1_2}), with almost no class separability. Furthermore, the final multimodal representations are essentially dominated by the audio modality (Fig. \ref{fig:class1_3}). In contrast, after applying the B-MM method, the performance of the audio modality shows almost no degradation (Fig. \ref{fig:class3_1}), but the feature space becomes significantly more compact, consistent with the intended effect of mixup. Meanwhile, the video modality shows a substantial improvement in separability compared to the baseline model (Fig. \ref{fig:class3_2}). These results further highlight the importance of B-MM in promoting modality balance during multimodal learning.

\subsection{Comparison with Balanced Multimodal Learning Methods}
In our mixup process, we take into account the modality imbalance problem in multimodal learning and design the Balanced Multimodal Mixup method based on the discrepancy ratio $\rho$ \cite{peng2022balanced}. To evaluate the advancement of our approach, we compare it with several representative methods that address multimodal balance and sufficiency, including G-Blend \cite{wang2020makes}, UMT \cite{du2021improving}, PMR \cite{fan2023pmr}, OGM-GE \cite{peng2022balanced}, Greedy \cite{greedy}, and ATF \cite{sun2021learning}. For fair comparison, we adopt concatenation as the baseline fusion strategy and follow standardized experimental settings commonly used in the field.

\begin{table*}[t]
\centering
  \caption{Ablation study on the effect of different fixed $\lambda$ values in the MM method on the CREMAD and Kinetic-Sounds datasets. ``(+)'' indicates performance improvement over the baseline, while ``(-)'' indicates performance degradation. The best result for each dataset is highlighted in bold.}
  \label{tbl:lambda}
  \begin{tabular}{ccccccc}
    \toprule
    Dataset & $\lambda = 0.05$& $\lambda = 0.1$ & $\lambda = 0.3$ & $\lambda = 0.5$ & $\lambda = 0.7$ & $\lambda = 0.9$ \\
    \midrule
    CREMAD & 61.96\% (+) & 63.17\% (+) & \textbf{64.65\%} (+) & 63.58\% (+) & 58.06\% (-) & 57.39\% (-) \\
    Kinetic-Sounds & 50.12\% (+) & \textbf{50.89\%} (+) & 49.04\% (+) & 49.11\% (+) & 48.46\% (-) & 47.42\% (-) \\
    \bottomrule
  \end{tabular}
\end{table*}

By examining the results in Table \ref{tab:conventional} and Table \ref{tab:bml}, we observe that applying our MM method effectively alleviates model overfitting and improves performance. However, as MM does not address the fact that overfitting mainly arises from the strong modality, its performance still lags behind that of balanced learning methods to some extent. In contrast, the B-MM method achieves greater performance gains, as it accounts for the differences in learning effectiveness across modalities and applies dynamic Mixup accordingly. As a result, B-MM significantly outperforms conventional balance learning approaches.

\subsection{Ablation Staudy}
We first conduct an ablation study on the $\lambda$ parameter in the Multimodal Mixup method to investigate its impact on model performance, as shown in Table \ref{tbl:lambda}. We observe that a very small or large value of $\lambda$ will result in performance degradation, indicating that excessive or insufficient mixing weakens the benefits of the Mixup strategy. 

These results suggest that an appropriate degree of mixing is crucial for balancing data augmentation and preserving the integrity of modal information, and highlight the necessity of adaptively tuning $\lambda$ for different tasks and datasets.

\begin{table*}[t]
\centering
  \caption{Ablation study on the effect of different numbers (n) of warm-up epochs before applying the B-MM method on the CREMAD and Kinetic-Sounds datasets. The best result for each dataset is highlighted in bold.}
  \label{tbl:epoch}
  \begin{tabular}{ccccccc}
    \toprule
    Dataset & $ n = 0$& $n=5$ & $n=10$ & $n=15$ & $n=20$ & $n=25$ \\
    \midrule
    CREMAD & 67.61\%& 67.61\%& \textbf{69.22\%}& 65.59\%& 65.46\%& 64.65\% \\
    Kinetic-Sounds & 52.70\%& \textbf{53.66\%}& 53.59\%& 53.20\%& 52.27\%& 52.69\% \\
    \bottomrule
  \end{tabular}
\end{table*}

\begin{table*}[t]
\centering
  \caption{Ablation study on the effect of the parameter $\alpha$ that controls the degree of Mixup. The best result for each dataset is highlighted in bold.}
  \label{tbl:alpha}
  \begin{tabular}{ccccccc}
    \toprule
    Dataset & $\alpha = 0.05$& $\alpha = 0.1$ & $\alpha = 0.3$ & $\alpha = 0.5$ & $\alpha = 0.7$ & $\alpha = 0.9$ \\
    \midrule
    CREMAD & 67.47\%& \textbf{69.22\%}& 68.15\%& 68.28\%& 67.34\%& 67.88\% \\
    Kinetic-Sounds & 51.92\%& 52.85\%& \textbf{53.66\%}& 52.46\%& 52.69\%& 51.65\% \\
    \bottomrule
  \end{tabular}
\end{table*}

Next, we conduct ablation studies on two key hyperparameters in the Balanced Multimodal Mixup method: the number of warm-up epochs before applying B-MM and the parameter that controls the degree of Mixup. The results on the CREMAD and Kinetic-Sounds datasets are presented in Table \ref{tbl:epoch} and Table \ref{tbl:alpha}, respectively. 

By examining the results, we observe that performing an appropriate warm-up phase before applying the B-MM method helps ensure that the model achieves a basic level of performance. This prevents the model from being exposed too early to complex virtual samples, which could otherwise hinder sufficient learning of each modality. In addition, the choice of $\alpha$ should not be too large, as there are inherent interactions between different modalities. A large $\alpha$ value would lead to overly aggressive mixing, preventing the model from adequately learning cross-modal mutual information \cite{han2021improving} and ultimately resulting in performance degradation.

\section{Conclusion}

In this work, we explored the challenge of modality imbalance in multimodal video understanding and proposed two complementary methods: Multimodal Mixup (MM) and Balanced Multimodal Mixup (B-MM). MM introduces mixup at the multimodal feature level to mitigate overfitting by enriching the training distribution with virtual feature-label pairs. Building on this foundation, B-MM further addresses the imbalance among modalities by dynamically adjusting mixing strategies based on each modality’s contribution during training. Extensive experiments on CREMAD, Kinetic-Sounds, and UCF-101 demonstrated that our methods consistently outperform conventional fusion strategies and existing balanced learning approaches, achieving better generalization and more robust multimodal cooperation.

Despite these promising results, several open questions remain. For instance, how can the dynamic Mixup strategy be extended to handle more than two modalities or to adapt to scenarios with missing or noisy modalities? Furthermore, integrating our approach with large-scale pretrained multimodal models and investigating its impact on tasks beyond classification, such as video captioning or temporal localization, are valuable directions for future research. We hope this work inspires further exploration of adaptive data augmentation for multimodal learning.


\nocite{}

\bibliography{main}
\bibliographystyle{icml2025}

\end{document}